\documentclass[runningheads]{llncs}

\usepackage{eccv}

\usepackage{eccvabbrv}

\usepackage{graphicx}
\usepackage{booktabs}
\usepackage{bm}
\usepackage{multicol}
\usepackage{multirow}
\usepackage{enumitem}
\usepackage{tabularx}

\usepackage[accsupp]{axessibility}  

\usepackage[pagebackref,breaklinks,colorlinks,citecolor=eccvblue]{hyperref}

\usepackage{orcidlink}
\usepackage{pdfpages}

\begin{document}

\title{Scene Graph Aided Radiology Report Generation} 

\titlerunning{Abbreviated paper title}

\author{Jun Wang\inst{1,2} \and 
Lixing Zhu\inst{2} \and
Abhir Bhalerao\inst{1} \and 
Yulan He\inst{1,2}}


\institute{University of Warwick, UK \and King's College London, UK
}

\maketitle

\begin{abstract}
   Radiology report generation (RRG) methods often lack sufficient medical knowledge to produce clinically accurate reports. The scene graph contains rich information to describe the objects in an image. We explore enriching the medical knowledge for RRG via a scene graph, which has not been done in the current RRG literature. To this end, we propose the Scene Graph aided RRG (SGRRG) network, a framework that generates region-level visual features, predicts anatomical attributes, and leverages an automatically generated scene graph, 
   thus achieving medical knowledge distillation in an end-to-end manner. SGRRG is composed of a dedicated scene graph encoder responsible for translating the scene graph, and a scene graph-aided decoder that takes advantage of both patch-level and region-level visual information. A fine-grained, sentence-level attention method is designed to better distill the scene graph information. 
   Extensive experiments demonstrate that SGRRG outperforms previous state-of-the-art methods in report generation and can better capture abnormal findings.\footnote{Code will be released upon the acceptance.}
\end{abstract}

\section{Introduction}
Interpreting radiographs, e.g., Chest X-rays, 
is of great importance in clinical practice. 
Therefore, RRG, aiming to automatically describe radiography in human language, has gained increasing attention because of its great potential to ease the burden on clinical resources and support the diagnostic process. Recently, some valuable insights and improvements have been reported 
driven by the availability of large-scale datasets and the development of new models \cite{chen2021cross,wang2022cross,wang2022camanet,tanida2023interactive}. However, the performance of RRG still falls short of being suitable 
for routine clinical deployment. One primary reason for this limitation is that RRG requires generating significantly 
\textit{more} text ($4$-$8$ sentences) compared to traditional captioning tasks to adequately describe the findings, and often it involves more sophisticated semantic and discourse relationships. The severe data biases present in commonly used datasets further aggravate the problem, with significantly fewer abnormal samples hindering the model from capturing effectively abnormal information. Furthermore, even in pathological cases, such abnormal regions typically represent only a small proportion of an image, leading to the majority of report sentences describing normal findings.

\begin{figure}
\vspace{-5pt}
\centering
\begin{minipage}{.5\textwidth}
  \centering
  \includegraphics[width=.8\linewidth]{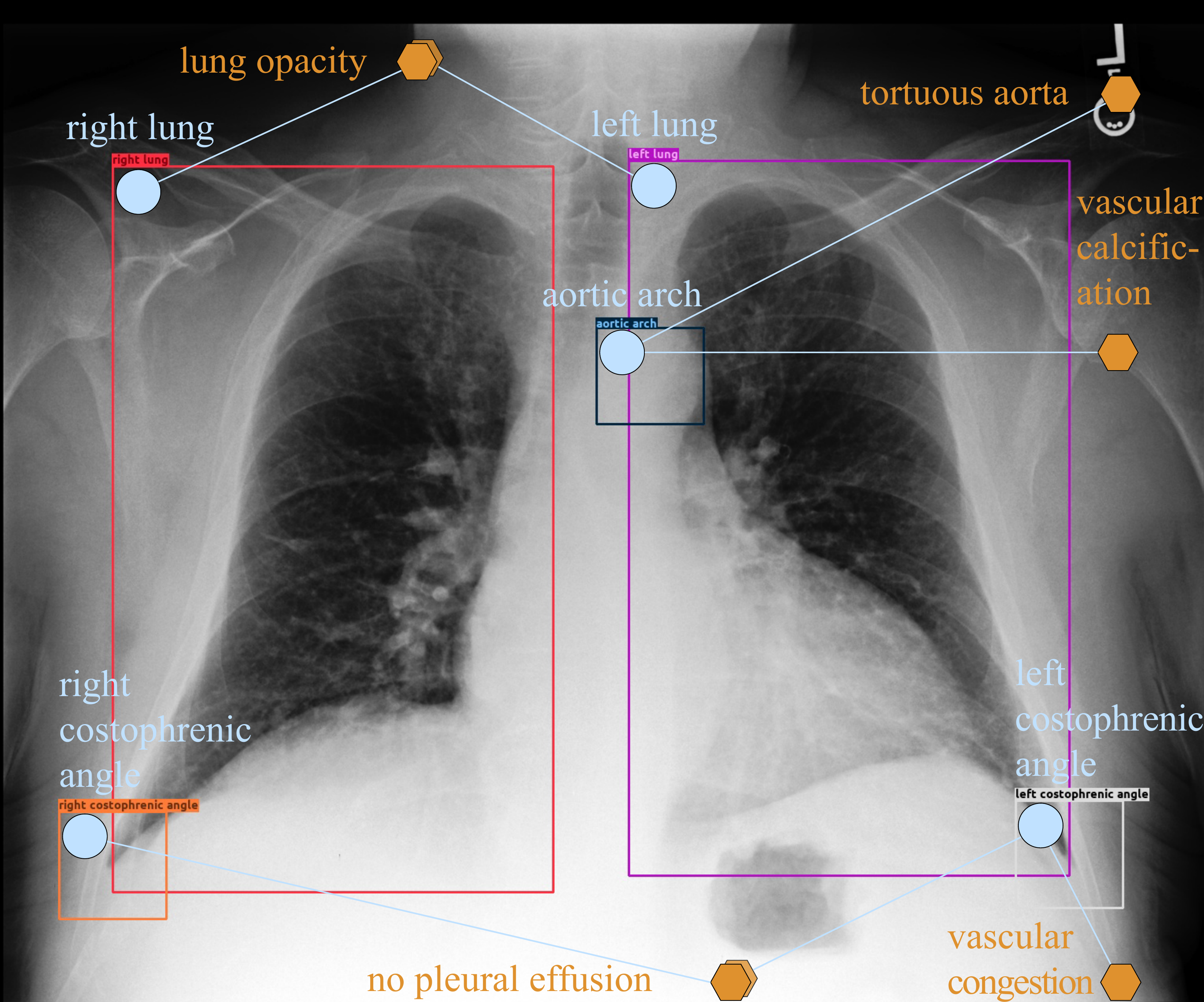}
  \captionsetup{width=.8\linewidth}
  \captionof{figure}{An example of radiology scene graph for parts of organs. The blue and red nodes refer to the object and attribute nodes.
  }
  \label{fig:scene_graph_intro}
\end{minipage}%
\begin{minipage}{.5\textwidth}
  \centering
  \includegraphics[width=.8\linewidth]{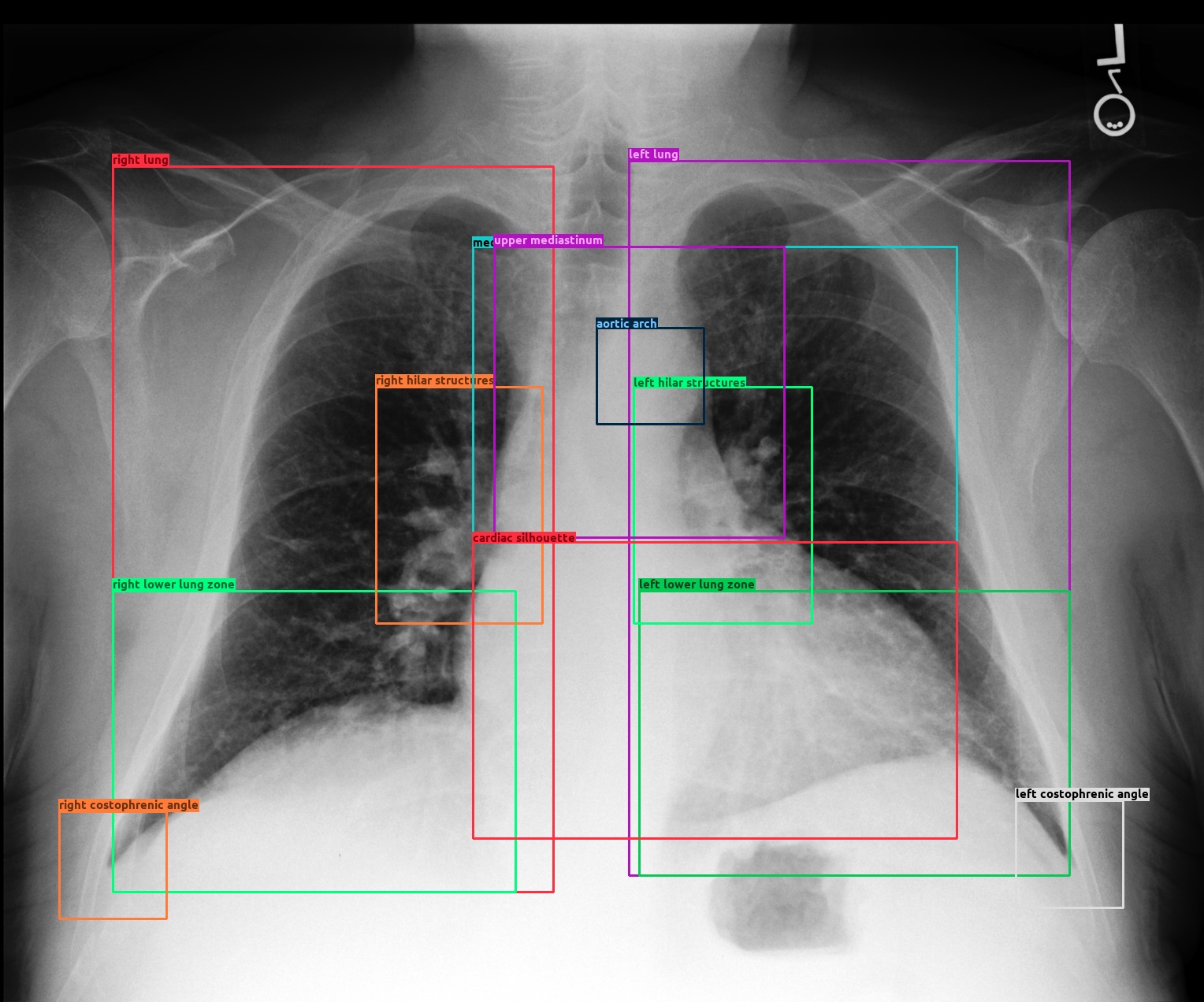}
  \captionsetup{width=.8\linewidth}
  \captionof{figure}{An image with bounding boxes showing the severe overlapping and noisy annotation problems.
  }
  \label{fig:overlapping_boxes}
\end{minipage}
\vspace{-30pt}
\end{figure}


RRG is a challenging task and therefore requires adequate medical knowledge to produce a clinically accurate report. Previous studies have typically incorporated medical knowledge from external sources such as other similar images or reports \cite{liu2021exploring,liu2021contrastive}, medical entities \cite{miura2021improving}, or a predefined knowledge graph~\cite{li2019knowledge,zhang2020radiology}. However, these approaches often overlook the inherent knowledge contained within each individual sample. To this end, this paper focuses more on exploring and harnessing the latent knowledge present within each sample. Scene graphs establish connections between objects and their associated attributes. As shown in \autoref{fig:scene_graph_intro}, they offer valuable source of rich contextual knowledge and high-level visual patterns. Leveraging such information is vital  
for writing accurate and comprehensive reports. Particularly,  
the radiology scene graphs used in this work are automatically generated based on both image and text, which may further introduce useful language inductive bias, thus bridging the gap between these two modalities.

Motivated by this, we propose a novel framework called Scene Graph aided RRG (SGRRG), which leverages the automatically generated scene graph with a transformer model to reveal latent knowledge within each sample, whilst preserving the important original patch-level visual patterns. Specifically, we translate the scene graph via a carefully designed scene graph encoder and introduce a fine-grained token type embedding method to address a severe problem of overlapping anatomical regions in radiology scene graphs, as illustrated in \autoref{fig:overlapping_boxes}. The encoded scene graph, combined with the visual and textual tokens, are then fed into a scene graph-aided decoder with a fine-grained distillation attention mechanism to distill the scene graph knowledge into the model. In addition, $SGRGG$ integrates key scene-graph generation processes into its generative model, including region-level representation generation, region selection, and attribute prediction. This design renders SGRRG a highly adaptable framework, enabling the easy incorporation of other advanced RRG techniques, such as patch-level representation learning or knowledge enrichment methods. Moreover, we introduce two abnormal information learning modules to facilitate the generation of clinically accurate reports. This work makes four primary contributions:
\begin{enumerate}
[noitemsep,nolistsep]
    \item A novel scene graph-aided framework, SGRRG, which effectively leverages the radiology scene graph and a transformer to extract and distill valuable knowledge present in each sample for the RRG task. To the best of our knowledge, this is the first work to explore using a radiology scene graph in this way for RRG.
    \item Our approach can better capture the abnormal information through the incorporation of two abnormal representation learning modules, applied to images features and scene graph representations, respectively.
    \item Our model benefits from both global and local information, and end-to-end training, owing to its carefully designed architecture.
    \item SGRRG shows promising results on the largest RRG benchmark MIMIC-CXR, surpassing multiple state-of-the-art methods and exhibiting superior capabilities in capturing abnormal findings.

\end{enumerate}

\section{Related Works}
\vspace{-5pt}
\subsection{Radiology Report Generation}
Recent RRG methods follow the same encoder-decoder architecture as an image captioning (IC) task, but a significant performance gap can be seen between these two tasks, mainly due to the aforementioned problems. Some research efforts \cite{jing2018automatic,xue2019improved} have been made to alleviate the problem of long text generation using the hierarchical LSTM network. Another group of works \cite{wang2022cross,chen2021cross} focus on improving the cross-modal interaction, 
while the most recent successful studies explore how to utilize medical domain knowledge to aid RRG. For example, Liu \etal~\cite{liu2021exploring} presents a model based on prior and posterior knowledge from other similar images/reports. Wang \etal~\cite{wang2022cross} and Yang \etal~\cite{yang86radiology} introduce disease tags to enrich abnormal and cross-modal knowledge. Zhang \etal~\cite{zhang2020radiology} and Kale \etal~\cite{kale2023kgvl} employ the predefined knowledge graph with the attention mechanism to generate descriptive reports. These knowledge-based approaches demonstrate improved performance on RRG, while most of them overlook the rich inherent knowledge that already exists within each sample itself. This paper therefore investigates how to leverage a radiology scene graph to extract such inherent knowledge within each sample and better aid the RRG task.

\vspace{-10pt}

\subsection{Scene Graph}
\vspace{-5pt}
Visual scene graphs \cite{krishna2017visual} typically comprise the knowledge of the objects present, their associated attributes and pairwise relationship among different objects, hence encapsulating the high-level semantic contents in an image. 
Several computer vision tasks,  e.g., image generation \cite{johnson2018image,zhao2019image} and visual question answering \cite{shi2019explainable,teney2017graph}, have successfully employed the scene graph and acheiving improved performance. Some studies explore it in captioning tasks where the earliest works \cite{yao2018exploring,yang2020auto,nguyen2021defense} employ a LSTM-GNN architecture to embed the scene graph into a captioning model. 
To profit from a transformer model \cite{vaswani2017attention}, one recent work \cite{yang2023transforming} form a homogeneous framework by exploiting the transformer with attention mechanism to encode and translate the scene graph into image captioning. These methods have brought improvements in image captioning, but however are not appropriate to RRG as they require visual sources from a well-trained and accurate region-level feature detector. Recently, radiology scene graphs \cite{wu2021chest}
were automatically established by rule-based natural language processing and a CXR atlas-based bounding box detection pipeline, leading to noisy and highly overlapped bounding box annotations in the training set.  Worse still, some radiographs do not predict any anatomical location instances (objects) at the inference phase by a detector trained under these annotations, making these region-only-based scene graph methods unreliable for RRG.

To this end, we propose a new RRG framework named SGRRG which leverages the scene graph to enrich medical knowledge, while benefiting from both the global-level and patch-level image features. SGRRG contains a variety of components, each tailored to RRG for accurate report generation. 

\begin{figure}[t]
\vspace{-10pt}
\centering
\centering
\includegraphics[width=0.9\textwidth]{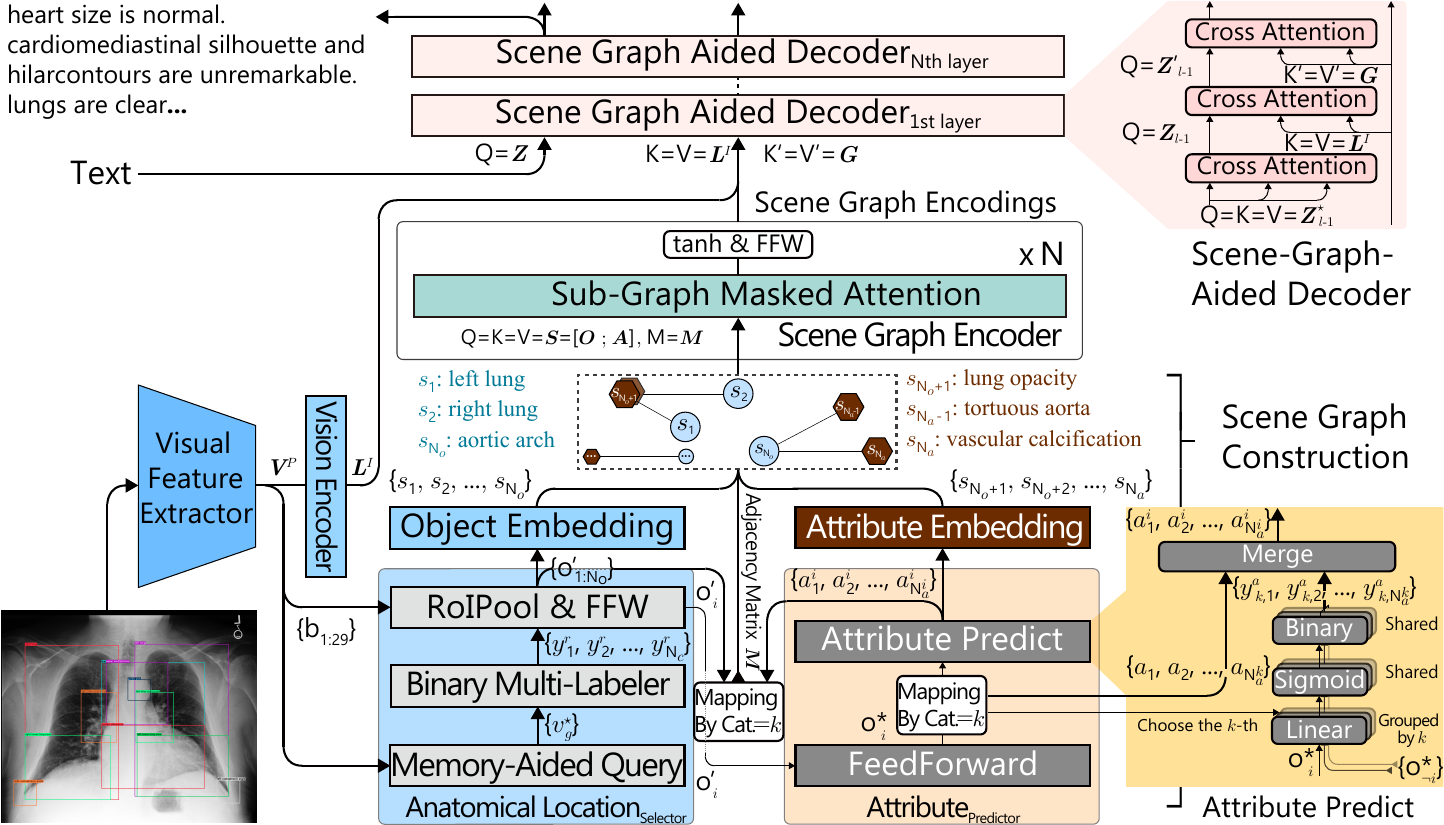}
\caption{The overall architecture of proposed $SGRRG$. The basic visual features obtained from the visual feature extractor are fed into the vision encoder, scene graph construction module and the scene graph encoder to obtain the encoded global and region level representations respectively. These representations are then taken as the input source for the proposed scene graph-aided decoder to predict the reports. }
\label{fig:s2-1}
\vspace{-20pt}
\end{figure}

\vspace{-10pt}
\section{Methodology}
As shown in Figure~\ref{fig:s2-1}, SGRRG comprises a visual feature extractor, a vision encoder, a scene graph construction module, a scene graph encoder, and a scene graph-aided decoder. Details of each module are given in the following sections.

\vspace{-5pt}
\subsection{Visual Feature Extractor and Vision Encoder}
Let a radiology image be denoted as $\bm{I}$, the purpose of RRG is to generate coherent findings (report) $\bm{R}=\{w_1,w_2,...,w_{N_r}\}$ from $\bm{I}$. $N_r$ refers to the number of words in the report. 
$\bm{I}$ is first sent to a visual feature extractor, e.g., Swin Transformer \cite{liu2021swin}, to extract the patch-level visual features $\bm{V}^p \in \mathbb{R}^{H \times W \times C}$ which are then flattened to a sequence of visual tokens $\bm{V}^s = \{ v^s_1, v^s_2,...,v^s_{N^s} \}$, where $v^s_k$ denotes the patch feature in the $k^{th}$ position in $\bm{V}^s$, and $N^s=H\times W$.

The vision encoder 
takes the extracted visual tokens as input to further capture the content and context/relationships in the image. Specifically, visual tokens are first projected into a latent space with dimension $D$ by a one-layer mapping network. 
These projected visual tokens are then fed into the vision encoder to perform the context modelling. Let the encoded visual tokens be denoted as $\bm{L}^I = \{l^I_1, l^I_2, ..., l^I_{N^s}\}$.

\subsection{Scene Graph Construction}
Generally, a visual scene graph contains three kinds of information: objects, attributes and relationships. These nodes are connected by the following rules: (1) an object $o_i$ is connected to an attribute $a_i$ if it has this attribute, and; (2) if two objects $o_i$ and $o_j$ have a relationship, then they are connected via $r_i$, denoted as $\langle o_i, r_i, o_j \rangle$. However, unlike a natural scene image, each sentence in the radiology report normally describes exactly one anatomy location (object). Hence, no explicit relationships among different objects are shown in the radiology images. The radiology scene graph therefore only contains two kinds of nodes: objects and their associated attributes. We use the automatically generated scene graph annotation from the training set to enable our proposed scene graph-aided report generation. Nevertheless, given that the scene graph is often not available during the inference phase, we incorporate a scene graph construction process into our model to generate it for the inference phase, thus creating a more adaptable, multi-task framework. 

\noindent \textbf{Anatomical Location Selector}
To generate the bounding box annotations for the inference phase, we train a detector \cite{zhang2022dino} to predict the possible bounding boxes for the $29$ predefined anatomical locations defined for a radiology image,  using the annotations of the training set. However, within the RRG domain, 
not every detected location will be described in the reports 
since radiologists may choose to ignore certain normal regions. Also, there is only one instance for each anatomical location in the RRG domain. To reduce spurious features, we propose to perform a multi-label binary classification task to predict whether an anatomical location should be included in the scene graph. In detail, a linear layer with a Sigmoid function is applied to the average-pooled visual feature $v_g = \frac{1}{N_s}\sum^{N_s}_{i=1} v^p_i$:
\begin{align}
\label{eq:region_predict}
  \{y^r_1, y^r_2, ..., y^r_i, ..., y^r_{N_c}\} = \Gamma_{\alpha}(\mathrm{Sigmoid}(\bm{W}^T_r \cdot v_g)),
\end{align} 
where $\bm{W}_r \in \mathbb{R}^{C\times N_c}$ is the learnable weight and $N_c$ denotes the total number of anatomical locations (object categories). $\Gamma_{\alpha}(\cdot)$ is a threshold function mapping inputs larger than $\alpha$ to $1$, and 0 otherwise.  $y^r_i \in \{0,1\}$, indicates the inclusion of $i^{th}$ anatomical location. Only anatomical locations detected by the detector 
with a $y^r_i=1$ are included in the scene graph. 

Despite the use of binary classification, 
filtering out the irrelevant anatomical locations purely based on the image is still difficult. Inspired by~\cite{chen2021cross} in learning the cross-modal patterns,
we design a memory-aided region selector by employing a memory matrix and performing a query-response process to distil the prior knowledge into the selector. In particular, we randomly initialize a learnable memory matrix $\bm{P} \in \mathbb{R}^{N_p \times d}$ that records the region selection knowledge across images. To distil the prior knowledge into the current image, we perform a query and response process that selects the top relevant memory slots (query) from the memory matrix and fuses them with the visual features (response). Specifically, we first project the global visual feature $v_g$ into the same latent space as memory matrix: $v'_g = \bm{W}^T_g \cdot v_g$. The similarity $u_i$ between each memory slot $p_i$ and $v'_g$ is calculated by \autoref{eq:cal_sim}. Then, the top $\gamma$ memory slots with the highest similarities are selected to generate the response $r$ for the image:
\begin{align}
\label{eq:cal_sim}
  u_i = \frac{v'_g\bm{W}_q\cdot p_i\bm{W}_p}{\sqrt{d}},& \\
  \label{eq:cal_response}
  r =\frac{1}{\gamma} \sum^{\gamma}_{i=1} w_i* (p_i\bm{W}_r), \quad w_i =& \frac{e^{u_i}}{\sum^{\gamma}_{j=1}e^{u_j}},
\end{align} 
where $p_i$ is the $i^{th}$ row of $\bm{P}$; $\bm{W}_q \in \mathbb{R}^{d\times d}$, $\bm{W}_p \in \mathbb{R}^{d\times d}$ and $\bm{W}_r \in \mathbb{R}^{d\times d}$ are three learnable weights. $w_i$ denotes the weight of the $i^{th}$ memory slot. After obtaining the response $r$, we fuse the global visual feature with the response by concatenating them with a lightweight, feed-forward module:
\begin{align}
\label{eq:fuse}
  v^{\star}_g = \mathrm{Dropout}(\sigma(\mathrm{Linear}([v'_g, r]))).
\end{align} 

\noindent where $\sigma$ is the $GELU$ activation. This prior-knowledge-enriched feature $v^{\star}_g$ is the one finally fed into the \autoref{eq:region_predict} to perform the anatomical location selection.

\noindent \textbf{Attribute Predictor}
With the anatomical locations on hand, the next step is to predict the attributes associated with them, which collaboratively form the final scene graph at the inference phase. Similarly, this is achieved by integrating a multi-label attribute classification task into the model. Since the number of attributes varies between different anatomical locations, each anatomical location is equipped with a separate classification head. We apply this classification to each selected instance of the anatomical location in the image. An instance representation is denoted as $o_i$ (details how it is obtained from the bounding-box annotation is given below), the prediction of its relevant attribute is 
:
\begin{align}
\label{eq:att_predict1}
 o^{\star}_i &=\mathrm{FeedForward}(o_i), \\
\label{eq:att_predict2}
\{y^{a}_{k,1}, y^{a}_{k,2}, ..., &y^a_{k,{N^k_a}}\} = \Gamma_{\beta}(\mathrm{Sigmoid}(o^{\star}_i \cdot \bm{W}^{k}_a)),
\end{align}
where $o_i$ is fed into a one-layer feed-forward module followed by the classification head to predict the presence ($y^a_{k,j} \in \{0,1\}$) of the $j^{th}$ attribute in anatomical location (category) $k=y(o_i)$. Here, $\bm{W}^{k}_a\in\mathbb{R}^{C\times N^{k}_a}$, and $N^k_a$ denote the weights and the number of attribute categories for the anatomical location $k$, respectively.

\subsection{Scene Graph Encoder}
This section presents the details of how we translate the radiology scene graph into the RRG model. 

\noindent \textbf{Object and Attribute Embedding} 
Given the bounding box and attribute annotations, we aim to generate dense object and attribute embeddings. Specifically, we apply the Region-of-Interest pooling \cite{girshick2015fast} to visual features $\bm{V}^p$ based on the bounding-box annotations to obtain the initial region-level object embedding $\bm{O}^{'} \in \mathbb{R}^{N_o \times C}$. Here, $N_o$ refers to the number of objects described in a radiology image. After this step, a one-layer feed-forward module 
is employed to improve the object representation by mapping it into the same latent space as the Vision Encoder and Decoder. The procedure to obtain the object embedding $\bm{O} \in \mathbb{R}^{N_o \times D}$ can be expressed as:
\begin{align}
\label{eq:object_embed}
  \{o'_1, o'_2, ..., o'_{N_o} \} &= \mathrm{RoIPool}(\bm{V}^p,\{b_1, b_2, ..., b_{N_o}\}), \\
  \label{eq:object_ff}
 \{o_1, o_2, ..., o_{N_o} \} &= \mathrm{FeedFoward}(\bm{O}^{'}),
\end{align}


To obtain the attribute embedding, we map each attribute to a unique index number through a mapping function $\mathrm{ID}(\cdot)$ and employ a trainable embedding layer to embed it into the same latent space as the object embedding. A layer normalization and a dropout layer are used to improve the robustness of the representation learning further:
\begin{align}
\label{eq:att_embed}
  a^{i*}_j =& \mathrm{Embed}(\mathrm{ID}(m^i_j)), \\
  \label{eq:att_ff}
 \{a^i_1, a^i_2, ..., a^i_{N^i_a} &\} = \mathrm{Drop}(\mathrm{Norm}(\bm{A}^{i*})), 
\end{align}

where $m^i_j$ and $a^i_j$ denote the $j^{th}$ attribute and embedded attribute for $i^{th}$ object in image $\bm{I}$ respectively, e.g., lung opacity for object right lung. $N^i_a$ is the number of attributes for the $i^{th}$ object. Note that $N_a^i$ differs depending on the type of object.


\noindent \textbf{Scene Graph Encoding} 
We employ an attention mechanism~\cite{vaswani2017attention} to gauge co-occurrences and learn the contextual knowledge of the embeddings. Akin to~\cite{yang2023transforming} who replaces a GNN with homogeneous Transformers, we adopt the Transformer architecture to encode the scene graph after obtaining the original embeddings. 
Specifically, given the object embeddings $\bm{O}$ and attribute embeddings $\bm{A}$, we first append attribute embeddings to object embeddings to form a sequence of node tokens. Then, a trainable token-type embedding is added to enable the encoder to recognize the type of tokens:
\begin{align}
\label{eq:obj_token}
  \textbf{Object}: \ \  &s_i = o_i + e_o,& &1\leq i\leq N_o, \\
  \label{eq:att_token}
  \textbf{Attribute}: \ \ &s_{N_o+j} = a_j + e_a,& &1\leq j\leq N_a,
\end{align}
The final node token set, $\bm{S}=\{s_1, s_2,...,s_{N_s}\}$, comprises both object embeddings and their associated attribute embeddings, where $N_s = N_o + N_a$.
However, as shown in \autoref{fig:overlapping_boxes}, objects in the radiology image are normally highly overlapped. 
This hinders the encoder's ability to discern distinct objects and impedes the learning of contextual knowledge. Possible noisy box annotations further exacerbate this problem. To tackle this, we propose applying a fine-grained token-type embedding, namely anatomy embedding. We add to the object embeddings with a learnable token-type embedding for each anatomical category. \autoref{eq:obj_token} then becomes \ref{eq:obj_token_fin} where $y(o_i)$ indexes the anatomical category for the $i^{th}$ object and $e^{y({o_i})}_{o}$ denotes the token type embedding for anatomical category $y(o_i)$.
\begin{align}
\label{eq:obj_token_fin}
  \textbf{Object}: \ \  s_i = o_i + e^{y(o_i)}_{o}, \quad 1\leq i\leq N_o,
\end{align}

After obtaining the initial node tokens, the next step is to encode the topology of the graph. We utilize the attention mask $\bm{M} \in \mathbb{R}^{N_s \times N_s}$ to depict the connectivity of nodes. In particular, we mark $\bm{M}_{i,j}=1$ if node $i$ and $j$ are connected directly in the original scene graph, otherwise $\bm{M}_{i,j}=0$. Note that $\bm{M}$ is a symmetric matrix, i.e., $\bm{M}_{i,j}=\bm{M}_{j,i}$, analogous to an adjacency matrix. The node tokens $\bm{S}$ together with this attention mask $\bm{M}$ are then fed into the scene graph encoder consisting of several transformer blocks to perform the scene graph encoding. Through the attention mask $\bm{M}$, each position of the higher Transformer layer mask-attend to the visible positions of the lower layer, acting as a graph aggregator,
 as expressed in \autoref{eq:mask_attn}. In this way, the scene-graph encoder can model the long-range dependency between connected nodes and capture the topology. Thus:
\begin{align}
\label{eq:mask_attn}
 \bm{V}^{\star}=\mathrm{Softmax}(\bm{M} \circledast \frac{\bm{Q}\bm{W}_q(\bm{K}\bm{W}_k)}{\sqrt{d}}) \bm{V}\bm{W}_v,
 \end{align} 
 where $\bm{Q}=\bm{K}=\bm{V}$ are the node tokens,
 $\bm{V}^{\star}$ denotes node tokens after masked-attention. $\circledast$ is the Hadamard product operation. 
 We denote the topology-enriched scene graph embeddings as $\bm{\mathcal{G}}$ output by the final layer of the encoder. 

\noindent \textbf{Scene Graph Aided Decoder}
Given the report $\bm{R}$, the encoded visual tokens $\bm{L}^I$ 
and the encoded scene graph $\bm{\mathcal{G}}$, we are able to train a mapping between them with a decoder which performs cross-modal fusion and autoregressively generates the report word by word. Here, we propose a transformer-based scene graph aided decoder to further distill the encoded scene graph $\bm{G}$ into the model with that cross-attend to the interleaved inputs.
In particular, an embedding layer is first created through tokenization to produce a sequence of dense representations of the report $\bm{E}^t=\{e^t_1,e^t_2,...,e^t_{N_r}\}$. Then, in each transformer block\footnote{we omit the $\mathrm{FeedForward}$ and $\mathrm{Add\&Norm}$ after each attention layer for clarity.}, we perform a self-attention within textual tokens $\bm{E}^t$, a cross-attention between textual tokens and visual tokens, and finally a cross-attention between the visual-enriched textual tokens and scene graph representation as expressed below:
\begin{align}
\label{eq:decode1}
 \bm{Z}_l =& \mathrm{Self\text{-}Attn}(\bm{Z}^{\star}_{l-1}, \bm{Z}^{\star}_{l-1}, \bm{Z}^{\star}_{l-1}), \\
\label{eq:decode2}
 \bm{Z}'_l &= \mathrm{Cross\text{-}Attn}(\bm{Z}_{l}, \bm{L}^I, \bm{L}^I), \\
 \label{eq:decode3}
 \bm{Z}^{\star}_l &= \mathrm{Cross\text{-}Attn}(\bm{Z}'_{l}, \bm{\mathcal{G}}, \bm{\mathcal{G}}), 
\end{align}
where $\bm{Z}^{\star}_l$ denotes the textual tokens output from the $l^{th}$ transformer layer and $\bm{Z}^{\star}_0 = \bm{E}^t$ is the initial embedded textual tokens. The textual tokens first interact with all visual tokens to absorb basic visual information, which may contain noisy and unimportant knowledge. Then, we further distill the scene graph $\bm{\mathcal{G}}$ encapsulating the essential visual information and context knowledge into the textual tokens. Through this coarse-to-fine scenario, the model benefits from global, local, and context knowledge.

As mentioned above, no explicit relationship is shown among different objects in the automatically generated scene graph. Hence, each scene graph actually consists of several sub-graphs which typically contain one anatomical location instance and its associated attributes, and corresponds to one sentence in the report. This observation inspired us to design a more fine-grained way to distill the scene graph. Intuitively, instead of requiring the model to find the important information from a scene graph representation containing both the object and attribute nodes from different anatomical locations for each textual token, it is relatively easier to learn useful information from a structured and refined representation, especially for RRG scene graph where each anatomical location instance generally has $2-5$ attributes and one scene graph may contain more than $80$ node tokens. To achieve this, we propose a sub-graph (sentence) level attention by utilizing  max-pooling to encapsulate the core knowledge in each sub-graph and then distilling them into the model:
\begin{align}
\label{eq:max_pool}
 &g'_i = \mathrm{MaxPool}(\{h^{\star}_i, h^i_1, ...,h^i_{N^i_a}\}), \\
\label{eq:sub_graph_attn}
 \bm{Z}^{\star}_l &= \mathrm{Cross\text{-}Attn}(\bm{Z}'_{l}, \{g'_1,...g'_{N_o}\}, \{g'_1,...g'_{N_o}\}), 
\end{align}
where $h^{\star}_i$, $h^i_j$ are the object tokens and its $j^{th}$ attribute token output by the scene graph encoder for $i^{th}$ anatomical location. Through the sub-graph attention, we explicitly alleviate the possible negative influence from other nodes (sentences) where the number of tokens from other sub-graphs (sentences) could reduce from more than $60$ to less than $20$. 

\vspace{-5pt}
\subsection{Abnormal Information Learning}
Information about abnormalities plays a vital role in generating clinically accurate reports. In this paper, we employ disease recognition and normal-abnormal segregation modules to enrich the abnormal information in the model.

\noindent \textbf{Disease Recognition}: As no ground-truth disease labels are available in the commonly used RRG benchmarks, we utilize pseudo-labels generated by an automatic labeller \cite{irvin2019chexpert} predicting 14 diseases on chest radiographs (shown in the supplementary file) to form a multilabel disease recognition task. In detail, we apply this task to the average-pooled visual features $v_g$ via a simple linear classification head, since the performance of the whole generative model is highly dependent on the quality of the extracted visual feature.

\noindent \textbf{Normal-Abnormal Segregation} (NAS): We also expect that our scene graph can better capture the abnormal information. To achieve this, we first leverage the attribute information to generate the abnormality label $y^d \in \{0,1\}$ for each sub-graph. All attributes belong to seven predefined types (listed in the supplementary file) and two of them could indicate the sample abnormality: (1) anatomical finding attribute with template \textit{`$\vert$\{yes,no\}$\vert$disease$\vert$'} such as \textit{`$\vert$yes$\vert$pneumothorax$\vert$'}; (2) NLP attribute with template \textit{`$\vert$\{yes,no\}$\vert$\{normal,} \textit{abnormal\}$\vert$'}. 
After obtaining the abnormality label, we use contrastive learning to improve abnormal scene graph representation learning. Note that, instead of classification, we choose contrastive learning as it is relatively soft and more appropriate for RRG given that the automatically generated scene graph may contain noisy and inaccurate supervision. We employ this contrastive learning to the global sub-graph representation $g'$ and \autoref{eq:con_ls} shows the calculation of the contrastive loss $L^k_{con}$ for the anatomical location $k$.
\begin{align}
\label{eq:con_ls}
\begin{split}
 L^k_{con} = \frac{1}{(N^k_b)^2}\sum^{N^k_b}_{i=1}\sum^{N^k_b}_{j:y^d_j=y^d_i}(1-\mathrm{Sim}(\epsilon(g'_i),\epsilon(g'_j))) + \\
 \sum^{N^k_b}_{j:y^d_j\not=y^d_i} \mathrm{max}(0,\mathrm{Sim}(\epsilon(g'_i),\epsilon(g'_j))-\omega),
 \end{split}
\end{align}
where $Sim$ and $\epsilon$ denote the cosine similarity and ${L2}$ normalization functions; $g'_i$ and $g'_j$ are the global sub-graph representation calculated by \autoref{eq:max_pool} from the same anatomical location $k$ in one batch; and $N^k_b$ is the number of sub-graphs from anatomical location $k$ in one batch and $\omega$ is the margin and set to $0.4$. By using this contrastive loss, we improve the scene graph abnormal representation learning by segregating the normal-abnormal information. Note that any contrastive loss can be applied here.

\vspace{-10pt}
\subsection{Objective Function}
Given the final textual tokens $\bm{Z_{L}}$ output by the decoder, a cross entropy loss $L_{gen}$ is used to supervise the report generation task, as shown in \autoref{eq:gen_loss}, where $\mathcal{C}$ and $y^r_i$ denotes the the classification head and the ground truth label for $i^{th}$ word in the report. Our model is jointly optimized by report generation loss $L_{gen}$, region selection loss $L_{rs}$, the attribute prediction loss $L_{ap}$, the disease recognition loss $L_{dr}$ and the normal-abnormal segregation loss $L_{con}$. The region selection and attribute prediction losses are implemented by the cross-entropy loss. \autoref{eq:fin_loss} shows the final loss $L_{fin}$: 
\begin{align}
\label{eq:gen_loss}
 L_{gen} &= -\frac{1}{N_r}\sum_{i=1}^{N^r}y^r_i\cdot \mathrm{log}(\mathrm{softmax}(\mathcal{C}(z_{l,i})), \\
 \label{eq:fin_loss}
L_{fin} &=L _{gen} +\lambda L_{rs} + \delta L_{ap} +  \eta L_{dr} + \varphi L_{con}
\end{align}
where $\lambda, \delta, \zeta$ and $\varphi$ are the hyper-parameters to control the contributions of different losses. One may ask why we do not directly use region features or the scene graph as the only visual source and remove the visual feature extractor. 
We address this question in a discussion in the supplementary files.

\begin{table}[t]
\caption{Comparative results of SGRRG with previous studies. The best values are highlighted in bold, and the second best are underlined. BL and RG, MTOR are the abbreviations of BLEU, ROUGE-L, and METEOR respectively.}
\vspace{-5pt}
\centering
\label{tab:main_results}
\small
\begin{tabular}{lcccccc|ccc}
\toprule  
\textbf{Method} & \textbf{BL-1} & \textbf{BL-2} & \textbf{BL-3} &
\textbf{BL-4} & \textbf{MTOR} &\textbf{RG-L} &\textbf{P} &\textbf{R} &\textbf{F1} \\
\midrule  
$PTSN$  &0.308 &0.196 &0.137 &0.102 &0.131 & 0.272 &0.285 &0.195 &0.193 \\ 
$\mathcal{M}^2_{NLL}$ &0.378 &0.247 &0.175 &0.132 &- &0.306 &\underline{0.422} &0.212 &0.234 \\ 
$FIBER$ &0.331 &0.207 &0.145 &0.108 &0.132 &0.266 &0.365 &0.225 &0.235 \\
$R2Gen$  &0.385 &0.251 &{0.178} &0.133 &0.151 & 0.313 &0.376 &0.203 &0.219 \\ 
$CMN$ &0.385 &0.252 &0.178 &0.133 &0.151 & 0.315 &0.357 &0.215 &0.218 \\
$M2KT$ &\underline{0.399} &0.240 &0.159 &0.112 &0.147 &0.289 &0.260 &0.195 &0.152 \\
$CAMANet$ &0.381 &0.249 &0.176 &0.131 &0.153 &0.315 &0.410 &0.244 &0.268 \\
$XProNet$ &0.394 &\underline{0.257} &\underline{0.182} &\underline{0.136} &0.161 & \underline{0.323} &0.360 &0.206 &0.209 \\
$RGRG$ &0.373 &0.249 &0.175 &{0.126} &\underline{0.168} & 0.264 & 0.380 &\underline{0.318} &\underline{0.305} \\
\bottomrule

$SGRRG$  &\textbf{0.434} &\textbf{0.285} &\textbf{0.202} &\textbf{0.149} &\textbf{0.173} & \textbf{0.326} & \textbf{0.440} &\textbf{0.323} &\textbf{0.332}\\

\bottomrule 
\end{tabular}
\vspace{-15pt}
\end{table}

\section{Experiments}
\subsection{Dataset and Evaluation Metrics}
We verify the effectiveness of our method in the Chest ImaGenome dataset \cite{wu2021chest} which is derived from the most widely used and largest RRG benchmark, i.e., MIMIC-CXR \cite{johnson2019mimic}, consisting of chest X-ray images with free-text reports. The Chest ImaGenome dataset provides automatically constructed scene graphs for the MIMIC-CXR frontal radiography where each scene graph contains bounding-box annotations for 29 unique anatomical locations, their associated attributes, and a label indicating if an anatomical location is described in the report. We adopt the official data split and follow previous works \cite{tanida2023interactive,wang2022cross} to discard reports with empty finding sections, resulting in 113,473 training images, 16498 validation images and 32711 test images.

We use six commonly used natural language processing metrics, i.e., \textit{BLEU}\{1-4\} \cite{papineni2002bleu}, \textit{Rouge-L} \cite{lin2004rouge} and \textit{METEOR} \cite{denkowski2011meteor}, to measure the performance of our model. To measure the model's capability of capturing the abnormalities, we follow previous works to report the clinical efficacy metric where CheXbert \cite{smit2020combining} is applied to labeling the generated reports and the results are compared with ground truth in 14 disease categories. We use macro-average precision, recall, and F1 score to measure clinical efficacy.

\subsection{Implementation details}
We adopt the Swin-B \cite{liu2021swin} pre-trained on ImageNet2K~\cite{deng2009imagenet} 
as our visual feature extractor. The vision encoder, scene-graph encoder, and scene-graph-aided decoder are randomly initialized comprising three transformer layers with a hidden size of $512$ and an attention head of $8$. The weights $\lambda$, $\delta$, $\eta$, $\varphi$ to control the loss contributions in \autoref{eq:fin_loss} are set to $0.25$, $0.1$, $0.25$, $0.1$. Further implementation details, e.g., data augmentation and learning rates, are given in the supplementary files.

\subsection{Comparison to SOTA methods}
Here, we compare our method with two representative image captioning models $PTSN$ \cite{zeng2022progressive} and $\mathcal{M}^2_{NLL}$ \cite{cornia2020meshed}, one recent pretrained vision-language model $FIBER$ \cite{dou2022coarse}, and recent state-of-the-art RRG studies including $R2Gen$ \cite{chen2020generating}, $CMN$ \cite{chen2021cross}, $XProNet$ \cite{wang2022cross}, $M2KT$ \cite{yang86radiology}, $CAMANet$ \cite{wang2024camanet} and $RGRG$ \cite{tanida2023interactive}. As shown in \autoref{tab:main_results}, $SGRRG$ achieves the best performance in all the NLP metrics, especially for BLEU score where our method surpasses the second-best performing method by a large margin (+ 3.5\% in B1 and +1.3\% in B4). In addition, $SGRRG$ also outperforms previous methods in all clinical efficacy metrics, demonstrating that it can better capture abnormal observations and produce clinically accurate reports. The method $RGRG$ takes the bounding boxes features as the only visual source, hence showing good performance in clinical efficacy than other previous methods, e.g., $R2Gen$, $XProNet$, and $CMN$ which consider the learnable patch-level visual features as input. However, since $RGRG$ loses important global information, it fails to produce the same coherent reports as other patch-level feature-based approaches, as indicated by NLP metrics. Nonetheless, $SGRRG$ obtains promising results on both the NLP and clinical efficacy metrics. We mainly attribute this to the rich context and domain knowledge in the scene graph and its bespoke architecture which takes the best advantage of both the global and local information with abnormal information enrichment. Note that the region-level features in $SGRRG$ are derived from the patch-level features, hence they are jointly optimised during the training, rather than fixed bounding box features that heavily rely on a specific detector and accurate annotations.

\begin{table}
\vspace{-15pt}
\caption{Ablation studies of each proposed component in $SGRRG$. \textbf{AVG.$\Delta$} refers to average improvement compared to the $Base$ model.}
\vspace{-5pt}
\setlength\tabcolsep{2.25pt}
\centering
\label{tab:ablaion_study}
\begin{tabular}{l|ccccc|cccc}
\toprule  
\textbf{Method}  & \textbf{BL1} & \textbf{BL4} &\textbf{MTR} &\textbf{RG} & \textbf{AVG.$\Delta$} &\textbf{P} &\textbf{R} &\textbf{F1} & \textbf{AVG.$\Delta$} \\
\midrule  
$Base$ &0.381 &0.134 &0.153 &0.318 &- &0.388 &0.241 &0.257 &-\\
$SGRRG$ &\textbf{0.434}  &\textbf{0.149} &\textbf{0.173} &\textbf{0.326} &\textbf{+$\bm{10.2}$\%} &\textbf{0.446} &\textbf{0.324} &\textbf{0.333} & \textbf{+$\bm{26.3}$\%} \\
\hline
w/o $SG$ &0.386  &0.135 &0.153 &0.317 &+$0.0$\% &0.420 &0.244 &0.262 &+$3.8$\%\\
w/o $SgAtt$  &0.427 &0.142 &0.170  &0.319 &+$7.4$\% &0.406 &0.327 &0.330 &+$22.9$\% \\
w/o $AE$  &0.412 &0.140 &0.166 &0.322 &+$5.6$\% &0.418 &0.321 &0.328 &+$22.9$\% \\
w/o $MEM$  &0.427 &0.142 &0.170 &0.318 &+$7.3$\% &0.439 &0.318 &0.336 &+$25.3$\% \\
w/o $NAS$  &0.418 &0.141 &0.167 &0.323 &+$6.4$\% & 0.410 &0.314 &0.320 &+$20.2$\%\\
w/o $DR$  &0.421 &0.141 &0.169 &0.323 &+$6.9$\% & 0.422 &0.282 &0.301 &+$14.3$\% \\

\bottomrule 

\end{tabular}
\vspace{-35pt}
\end{table}

\subsection{Ablation Studies}
\noindent \textbf{Contribution of each component} We conduct an ablation study to investigate the contribution of each proposed core module in SGRRG. Specifically, the $Base$ model consists of the visual extractor and the transformer as the vision encoder and decoder. We verify the performance improvements brought by six components: (1) \textbf{SG}: The whole scene graph design. (2) \textbf{SgAtt}: The Subgraph-level ATTention (3) \textbf{DR}: The disease recognition; (4) \textbf{NAS}: The Normal-Abnormal Segregation. (5) \textbf{AE}: The fine-grained Anatomy Embedding. (6) \textbf{MEM}: The MEMory mechanism in region selection.

As presented in \autoref{tab:ablaion_study}, $SGRRG$ improves remarkably the performance of $Base$ model on both the NLP and clinical efficacy metrics (+10.2\% and +26.3\% on average over all NLP and clinical efficacy metrics, respectively). Furthermore, each component of SGRRG can be found to have a positive effect on overall performance. Specifically, a significant performance drop can be seen when we remove the whole scene graph design meaning that current model resorts to $Base+DR$. These results verify the effectiveness of our scene-graph-aided design and confirm the theory of leveraging the radiology scene graph as a knowledge source. In addition, removing the $SgAtt$, $FTE$ and $MEM$ notably impedes the model performance on NLP metrics, while showing relatively modest influence on the clinical efficacy since these designs focus more on improving the scene graph representation. A more obvious performance drop is reported when the $AE$ module is removed, indicating the importance of alleviating the overlapping issues in noisy radiology scene graphs. Furthermore, segregating abnormal-normal sub-graphs and enriching visual features with disease information play an essential role in model performance, especially for clinically accurate report generation, as evidenced by comparing the last two rows with $SGRRG$.

\begin{table}
\vspace{-20pt}
\caption{Ablation studies of the pooling methods in sub-graph attention and normal-abnormal segregation modules.}
\setlength\tabcolsep{2.25pt}
\centering
\label{tab:pooling_ablaion}
\begin{tabular}{l|cccc|ccc}
\toprule  
\textbf{Method}  & \textbf{BL1} & \textbf{BL4} &\textbf{MTR} &\textbf{RG} & \textbf{P} &\textbf{R} &\textbf{F1} \\
\midrule  
$Average$ &0.421 &0.144 &0.169 &0.325 &0.440 &0.324 &0.323 \\
$Maximum$ &0.434  &0.149 &0.173 &0.326 &0.446 &0.324 &0.333 \\

\bottomrule 

\end{tabular}
\vspace{-35pt}
\end{table}

\subsubsection{Average Pooling vs Maximum Pooling}: We apply the maximum pooling in our sub-graph level attention and normal-abnormal segregation modules to obtain sentence-level representation. Here, we conduct an ablation study to explore the effect of average pooling and maximum pooling. It can be found from \autoref{tab:pooling_ablaion} that $SGRRG$ with maximum pooling obtains better performance in both the NLP and clinical efficacy metrics. This is expected as the $maximum$ pooling highlights the salient parts in a sub-graph which is more suitable for fine-grained distillation modules and abnormal information learning, while the average pooling may diminish the contribution for those salient parts. Further ablation studies, e.g., the capacity of the memory matrix and the margin in the NAS module, are shown in the supplementary files.

\begin{figure}
\vspace{-20pt}
\centering
\includegraphics[width=0.95\textwidth]{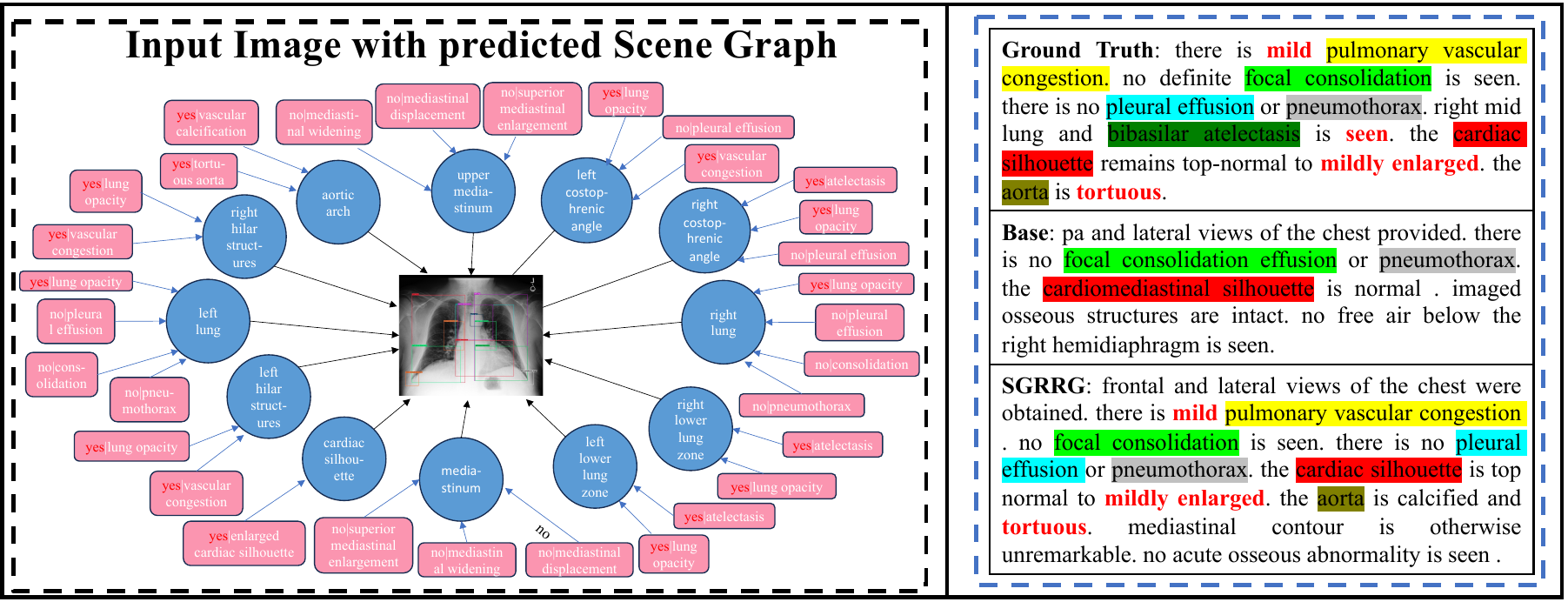}
\caption{An example of reports generated by different models. Most medical terms are highlighted. Abnormal terms are in bold.}
\label{fig:example_vis}
\vspace{-20pt}
\end{figure}

\subsection{Qualitative Results}
We present qualitative results to further demonstrate the effectiveness of $SGRRG$. As illustrated in \autoref{fig:example_vis}, $SGRRG$ appears to better capture disease information and generate abnormal descriptions, e.g., \textit{``there is mild pulmonary vascular congestion, the cardiac silhouette is top normal to mildly enlarged and the aorta is calcified and tortuous.''}, compared to the $Base$ model. To investigate whether this improvement is truly related to the radiology scene graph, we also visualize the predicted scene graphs with anatomical finding attributes in which the abnormal observations, e.g., \textit{``lung opacity", ``atelectasis", ``vascular congestion" and ``tortuous aorta"}, are present. Clearly, such abnormal medical observations are vital in report generation. Moreover, the average number of words in reports generated by $SGRRG$ is $14.5$\% longer than the base model, i.e., $50.5$ vs $44.1$ ($53.6$ in ground truth reports), indicating that $SGRRG$ could generate more descriptive reports. These results further verify the potential of the radiology scene graph and our proposed framework. Further examples are given in the supplementary files.

\vspace{-5pt}
\subsection{Limitations}
 Although we incorporate the core scene graph generation modules into the model to generate the scene graphs purely based on the image in the inference phase where reports are unavailable for scene graph mining required by recent automatic tools in radiology domain, our framework still depends on automatic tools to extract the scene graphs from the training set during training. Moreover, to attain enhanced performance, SGRRG necessitates an ancillary phase encompassing the construction and encoding of the scene graphs, and the abnormal information enrichment, which consequently prolongs the training time by $65.7$\% per epoch. While the increment in testing is small (+$19.4$\%), mainly attributed to the elimination of additional gradient computations associated with the losses from graph construction and abnormal information enrichment during the inference phase. Further code optimization may diminish the extra time cost. 

\vspace{-5pt}
\section{Conclusion}
We propose a novel RRG framework $SGRRG$ that harnesses the inherent knowledge within each sample by transforming the radiology scene graph into the model. A scene graph encoder is designed to encode the scene graph and alleviates the overlapping problem shown in the radiology domain. We construct a scene graph-aided decoder to distill the encoded scene graph into the model. $SGRRG$ integrates the scene graph generation into the model and takes advantage of visual patterns at the patch and region levels, forming a more flexible architecture through carefully designed modules. Extensive experiments verify the superior performance of $SGRRG$ and validate the effectiveness of each proposed module. This work also demonstrates the great potential knowledge that exists within each sample and presents an effective approach to leverage radiology scene graphs as an in-sample knowledge source.


%
%
\bibliographystyle{splncs04}
\bibliography{main}

\begin{thebibliography}{10}
\providecommand{\url}[1]{\texttt{#1}}
\providecommand{\urlprefix}{URL }
\providecommand{\doi}[1]{https://doi.org/#1}

\bibitem{chen2021cross}
Chen, Z., Shen, Y., Song, Y., Wan, X.: Cross-modal memory networks for radiology report generation. In: Proceedings of the 59th Annual Meeting of the Association for Computational Linguistics and the 11th International Joint Conference on Natural Language Processing (Volume 1: Long Papers). pp. 5904--5914 (2021)

\bibitem{chen2020generating}
Chen, Z., Song, Y., Chang, T.H., Wan, X.: Generating radiology reports via memory-driven transformer. In: Proceedings of the 2020 Conference on Empirical Methods in Natural Language Processing (EMNLP). pp. 1439--1449 (2020)

\bibitem{cornia2020meshed}
Cornia, M., Stefanini, M., Baraldi, L., Cucchiara, R.: Meshed-memory transformer for image captioning. In: Proceedings of the IEEE Conference on Computer Vision and Pattern Recognition. pp. 10578--10587 (2020)

\bibitem{deng2009imagenet}
Deng, J., Dong, W., Socher, R., Li, L.J., Li, K., Fei-Fei, L.: Imagenet: A large-scale hierarchical image database. In: 2009 IEEE Conference on Computer Vision and Pattern Recognition. pp. 248--255. Ieee (2009)

\bibitem{denkowski2011meteor}
Denkowski, M., Lavie, A.: Meteor 1.3: Automatic metric for reliable optimization and evaluation of machine translation systems. In: Proceedings of the sixth workshop on Statistical Machine Translation. pp. 85--91 (2011)

\bibitem{dou2022coarse}
Dou, Z.Y., Kamath, A., Gan, Z., Zhang, P., Wang, J., Li, L., Liu, Z., Liu, C., LeCun, Y., Peng, N., et~al.: Coarse-to-fine vision-language pre-training with fusion in the backbone. Advances in neural information processing systems  \textbf{35},  32942--32956 (2022)

\bibitem{girshick2015fast}
Girshick, R.: Fast r-cnn. In: Proceedings of the IEEE International Conference on Computer Vision. pp. 1440--1448 (2015)

\bibitem{irvin2019chexpert}
Irvin, J., Rajpurkar, P., Ko, M., Yu, Y., Ciurea-Ilcus, S., Chute, C., Marklund, H., Haghgoo, B., Ball, R., Shpanskaya, K., et~al.: Chexpert: A large chest radiograph dataset with uncertainty labels and expert comparison. In: Proceedings of the AAAI conference on artificial intelligence. vol.~33, pp. 590--597 (2019)

\bibitem{jing2018automatic}
Jing, B., Xie, P., Xing, E.: On the automatic generation of medical imaging reports. In: Proceedings of the 56th Annual Meeting of the Association for Computational Linguistics (Volume 1: Long Papers). pp. 2577--2586 (2018)

\bibitem{johnson2019mimic}
Johnson, A.E., Pollard, T.J., Greenbaum, N.R., Lungren, M.P., Deng, C.y., Peng, Y., Lu, Z., Mark, R.G., Berkowitz, S.J., Horng, S.: Mimic-cxr-jpg, a large publicly available database of labeled chest radiographs. arXiv preprint arXiv:1901.07042  (2019)

\bibitem{johnson2018image}
Johnson, J., Gupta, A., Fei-Fei, L.: Image generation from scene graphs. In: Proceedings of the IEEE conference on Computer Vision and Pattern Recognition. pp. 1219--1228 (2018)

\bibitem{kale2023kgvl}
Kale, K., Bhattacharyya, P., Gune, M., Shetty, A., Lawyer, R.: Kgvl-bart: Knowledge graph augmented visual language bart for radiology report generation. In: Proceedings of the 17th Conference of the European Chapter of the Association for Computational Linguistics. pp. 3393--3403 (2023)

\bibitem{krishna2017visual}
Krishna, R., Zhu, Y., Groth, O., Johnson, J., Hata, K., Kravitz, J., Chen, S., Kalantidis, Y., Li, L.J., Shamma, D.A., et~al.: Visual genome: Connecting language and vision using crowdsourced dense image annotations. International journal of computer vision  \textbf{123},  32--73 (2017)

\bibitem{li2019knowledge}
Li, C.Y., Liang, X., Hu, Z., Xing, E.P.: Knowledge-driven encode, retrieve, paraphrase for medical image report generation. In: Proceedings of the AAAI Conference on Artificial Intelligence. vol.~33, pp. 6666--6673 (2019)

\bibitem{lin2004rouge}
Lin, C.Y.: Rouge: A package for automatic evaluation of summaries. In: Text Summarization Branches Out. pp. 74--81 (2004)

\bibitem{liu2021exploring}
Liu, F., Wu, X., Ge, S., Fan, W., Zou, Y.: Exploring and distilling posterior and prior knowledge for radiology report generation. In: Proceedings of the IEEE Conference on Computer Vision and Pattern Recognition. pp. 13753--13762 (2021)

\bibitem{liu2021contrastive}
Liu, F., Yin, C., Wu, X., Ge, S., Zhang, P., Sun, X.: Contrastive attention for automatic chest x-ray report generation. In: Findings of the Association for Computational Linguistics: ACL-IJCNLP 2021. pp. 269--280 (2021)

\bibitem{liu2021swin}
Liu, Z., Lin, Y., Cao, Y., Hu, H., Wei, Y., Zhang, Z., Lin, S., Guo, B.: Swin transformer: Hierarchical vision transformer using shifted windows. In: Proceedings of the IEEE international Conference on Computer Vision. pp. 10012--10022 (2021)

\bibitem{miura2021improving}
Miura, Y., Zhang, Y., Tsai, E., Langlotz, C., Jurafsky, D.: Improving factual completeness and consistency of image-to-text radiology report generation. In: Proceedings of the 2021 Conference of the North American Chapter of the Association for Computational Linguistics: Human Language Technologies. pp. 5288--5304 (2021)

\bibitem{nguyen2021defense}
Nguyen, K., Tripathi, S., Du, B., Guha, T., Nguyen, T.Q.: In defense of scene graphs for image captioning. In: Proceedings of the IEEE International Conference on Computer Vision. pp. 1407--1416 (2021)

\bibitem{papineni2002bleu}
Papineni, K., Roukos, S., Ward, T., Zhu, W.J.: Bleu: a method for automatic evaluation of machine translation. In: Proceedings of the 40th annual meeting of the Association for Computational Linguistics. pp. 311--318 (2002)

\bibitem{shi2019explainable}
Shi, J., Zhang, H., Li, J.: Explainable and explicit visual reasoning over scene graphs. In: Proceedings of the IEEE conference on Computer Vision and Pattern Recognition. pp. 8376--8384 (2019)

\bibitem{smit2020combining}
Smit, A., Jain, S., Rajpurkar, P., Pareek, A., Ng, A.Y., Lungren, M.: Combining automatic labelers and expert annotations for accurate radiology report labeling using bert. In: Proceedings of the 2020 Conference on Empirical Methods in Natural Language Processing (EMNLP). pp. 1500--1519 (2020)

\bibitem{tanida2023interactive}
Tanida, T., M{\"u}ller, P., Kaissis, G., Rueckert, D.: Interactive and explainable region-guided radiology report generation. In: Proceedings of the IEEE Conference on Computer Vision and Pattern Recognition. pp. 7433--7442 (2023)

\bibitem{teney2017graph}
Teney, D., Liu, L., van Den~Hengel, A.: Graph-structured representations for visual question answering. In: Proceedings of the IEEE conference on Computer Vision and Pattern Recognition. pp.~1--9 (2017)

\bibitem{vaswani2017attention}
Vaswani, A., Shazeer, N., Parmar, N., Uszkoreit, J., Jones, L., Gomez, A.N., Kaiser, {\L}., Polosukhin, I.: Attention is all you need. Advances in Neural Information Processing Systems  \textbf{30} (2017)

\bibitem{wang2022cross}
Wang, J., Bhalerao, A., He, Y.: Cross-modal prototype driven network for radiology report generation. In: Computer Vision--ECCV 2022: 17th European Conference, Tel Aviv, Israel, October 23--27, 2022, Proceedings, Part XXXV. pp. 563--579. Springer (2022)

\bibitem{wang2022camanet}
Wang, J., Bhalerao, A., Yin, T., See, S., He, Y.: Camanet: Class activation map guided attention network for radiology report generation. arXiv preprint arXiv:2211.01412  (2022)

\bibitem{wang2024camanet}
Wang, J., Bhalerao, A., Yin, T., See, S., He, Y.: Camanet: class activation map guided attention network for radiology report generation. IEEE Journal of Biomedical and Health Informatics  (2024)

\bibitem{wu2021chest}
Wu, J., Agu, N., Lourentzou, I., Sharma, A., Paguio, J.A., Yao, J.S., Dee, E.C., Kashyap, S., Giovannini, A., Celi, L.A., et~al.: Chest imagenome dataset for clinical reasoning. In: Annual Conference on Neural Information Processing Systems (2021)

\bibitem{xue2019improved}
Xue, Y., Huang, X.: Improved disease classification in chest x-rays with transferred features from report generation. In: Information Processing in Medical Imaging: 26th International Conference, IPMI 2019, Hong Kong, China, June 2--7, 2019, Proceedings 26. pp. 125--138. Springer (2019)

\bibitem{yang86radiology}
Yang, S., Wu, X., Ge, S., Zheng, Z., Zhou, S.K., Xiao, L.: Radiology report generation with a learned knowledge base and multi-modal alignment. Medical image analysis  \textbf{86},  102798 (2023)

\bibitem{yang2023transforming}
Yang, X., Peng, J., Wang, Z., Xu, H., Ye, Q., Li, C., Yan, M., Huang, F., Li, Z., Zhang, Y.: Transforming visual scene graphs to image captions. In: Proceedings of the 61th Annual Meeting of the Association for Computational Linguistics (Volume 1: Long Papers) (2023)

\bibitem{yang2020auto}
Yang, X., Zhang, H., Cai, J.: Auto-encoding and distilling scene graphs for image captioning. IEEE Transactions on Pattern Analysis and Machine Intelligence  \textbf{44}(5),  2313--2327 (2020)

\bibitem{yao2018exploring}
Yao, T., Pan, Y., Li, Y., Mei, T.: Exploring visual relationship for image captioning. In: Proceedings of the European conference on computer vision (ECCV). pp. 684--699 (2018)

\bibitem{zeng2022progressive}
Zeng, P., Zhu, J., Song, J., Gao, L.: Progressive tree-structured prototype network for end-to-end image captioning. In: Proceedings of the 30th ACM International Conference on Multimedia. pp. 5210--5218 (2022)

\bibitem{zhang2022dino}
Zhang, H., Li, F., Liu, S., Zhang, L., Su, H., Zhu, J., Ni, L., Shum, H.Y.: Dino: Detr with improved denoising anchor boxes for end-to-end object detection. In: The Eleventh International Conference on Learning Representations (2022)

\bibitem{zhang2020radiology}
Zhang, Y., Wang, X., Xu, Z., Yu, Q., Yuille, A., Xu, D.: When radiology report generation meets knowledge graph. In: Proceedings of the AAAI Conference on Artificial Intelligence. vol.~34, pp. 12910--12917 (2020)

\bibitem{zhao2019image}
Zhao, B., Meng, L., Yin, W., Sigal, L.: Image generation from layout. In: Proceedings of the IEEE Conference on Computer Vision and Pattern Recognition. pp. 8584--8593 (2019)

\end{thebibliography}

\appendix



\section*{Supplementary material}

# Scene Graph Aided Radiology Report Generation - Appendix

**Author Name**
Affiliation
email@example.com

Table A1: The 14 disease categories predefined and output by the automatic labeler.

| Lung Opacity | Cardiomegaly | No Finding |
| Lung Lesion | Consolidation | Edema |
| Pneumothorax | Pneumonia | Atelectasis |
| Pleural Effusion | Pleural Other | Fracture |
| Support Devices | Enlarged Cardiomediastinum | |

## A.1 14 Categories of Chest X-Ray Observations

Table A1 shows the 14 disease categories for the disease recognition module set by the automatic labeller.

## A.2 Further Implementation Details

The radiologists can choose if include a normal regions into the report or not, which could, however, confuse the model during training. Therefore, we set a low threshold $\alpha = 0.2$ to include more possible anatomical locations and let the model learn how to select the relevant subgraphs. In addition, this also enriches the local information in the model. $\beta$ is set to the common threshold value $0.5$. For attribute predictions, we remove the attribute with occurrence less than $10$, resulting in a total of $849$ attributes for all the anatomical locations. The number of attributes for each anatomical location is listed in Table A2. The margin $\omega$ for the contrastive loss in Normal-Abnormal Segregation module is set to $0.4$.

Images are first cropped (resized) to $(224, 224)$ during the training (and inference) phase. In addition, we apply a random rotation to further augment the datasets during training. The downsampling rate of the visual extractor is $32$, hence there are $N^s = 7 \times 7 = 49$ visual tokens. AdamW [2] is used to optimise the whole model. The learning rates are set to $1e-4$ for the visual feature extractor and $2e-4$ for the remaining parts. We train our model with a batch size of $128$.

Following recent SOTA studies [1; 5; 4], we adopt Beam Search as the sampling method during the inference phase with the beam size of 3 to balance effectiveness and efficiency. We implement our model through the PyTorch [3] deep learning framework on Nvidia A100 GPU cards.

Table A2: The number of attribute (#.Attributes) for each anatomical location (Anatomy.Loc).

| Anatomy.Loc | #.Attributes | Anatomy.Loc | #.Attributes |
| --- | --- | --- | --- |
| *right lung* | 67 | *left lung* | 67 |
| *right clavicle* | 15 | *left clavicle* | 15 |
| *right apical zone* | 22 | *left apical zone* | 21 |
| *mediastinum* | 39 | *abdomen* | 9 |
| *spine* | 12 | *aortic arch* | 12 |
| *right atrium* | 7 | *trachea* | 10 |
| *carina* | 4 | *svc* | 9 |
| *right hemidiaphragm* | 13 | *left hemidiaphragm* | 12 |
| *right costophrenic angle* | 32 | *left costophrenic angle* | 32 |
| *right mid lung zone* | 47 | *left mid lung zone* | 45 |
| *right upper lung zone* | 46 | *left upper lung zone* | 43 |
| *right lower lung zone* | 56 | *left lower lung zone* | 57 |
| *right hilar structures* | 46 | *left hilar structures* | 46 |
| *upper mediastinum* | 28 | *cavoatrial junction* | 4 |
| *cardiac silhouette* | 33 | - | - |

## A.3 Attributes in Chest ImaGenome

The attributes in Chest ImaGenome give the descriptions that are true for different anatomical structures visualized on the CXR image, e.g., there is a right upper lung (object) opacity (attribute). Note that these attributes are automatically extracted by NLP tools without manual annotations. The attributes follow the pattern of $\langle categoryID|relation|label\_name\rangle$ (an attribute instance can be "*anatomicalfinding|no|lung opacity*"), where categoryID is the radiology semantic category the authors of Chest ImaGenome gave to the CXR concept in consultation with multiple radiologists, and relation is the NLP context relating the $label\_name$ to the anatomical location. If the relation is 'no', then the $label\_name$ is specifically negated in the sentence. If the relation is 'yes', then the $label\_name$ is affirmed in the sentence. All attributed belong to 7 categories:

1. **Anatomical Findings**: describes findings of anatomies where there is some subjectivity in the grouping of the phrases used to extract the labels.
2. **NLP**: normal or abnormal descriptions about different anatomical locations and can be somewhat subjective.
3. **Disease**: descriptions that are more diagnostic level

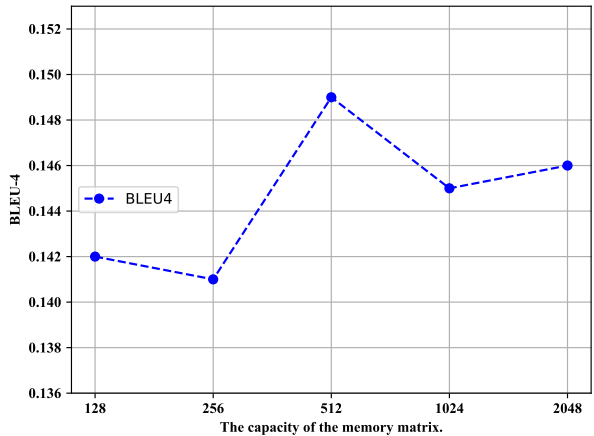

Figure A1: Ablation study for the capacity of the memory matrix.

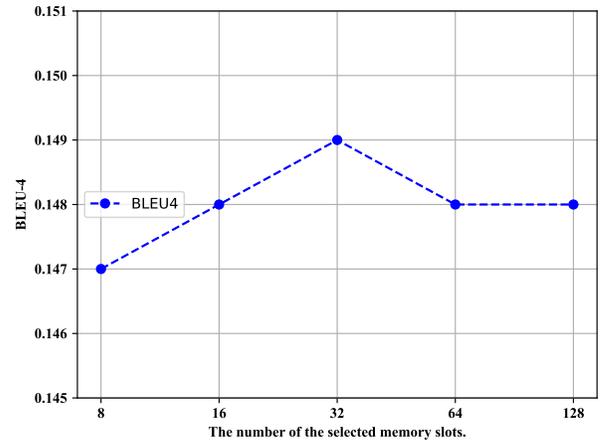

Figure A2: Ablation study for the number of selected memory slots in anatomical location selector.

Table A3: The accuracy of anatomical location selection for different methods.

4. **Technical Assessment**: image quality issues that affect radiologic interpretation of imaging observations.
5. **Tubes and Lines**: medical support devices where radiologists need to report any placement issues.
6. **Devices**: medical devices where placement issues are less relevant.
7. **Texture**: these are only present in the "texture_cues" field. The author of Chest ImaGenome kept a set of highly non-specific attributes (e.g. opacity, lucency, interstitial, airspace) that tend to form the initial most objective descriptions about what is observed in the images by radiologists.

## A.4 Further Ablation Studies

**The influence of the capacity of the memory matrix**: we investigate the influence of the capacity of the memory matrix $N_p$, i.e., the total number of memory slots in the memory matrix, by varying the $N_p$ from 128 to 2048 and fixing the number of selected slots to 32. As Figure A1 shows, $SGRRG$ is not overly sensitive to the capacity of the memory matrix. Nonetheless, selecting an appropriate capacity is of greate importance, since a small memory matrix cannot learn and record sufficient patterns for anatomical location selection while a too large $N_k$ increases the difficulty in learning the most related prior knowledge from a large memory matrix.

**The influence of the number of selected memory slots**: we also investigate $\gamma$, the influence of the number of selected memory slots to perform the response operation, in the memory-aided anatomical location selector. Specifically, we fix the number of total memory slots to 512 and vary $\gamma$ from 8 to 128. As shown in Figure A2, $SGRRG$ is not sensitive to the number of selected memory slots. Increasing $\gamma$ allows the model to interact with more patterns and prior knowledge from other samples, while a too large $\gamma$ could introduce noisy information.

**Anatomical Location Selector Accuracy**: We also conduct an ablation study to further verify whether the memory mechanism improves the anatomical location selection performance. We use the Area Under the Receiver Operating Characteristic Curve (AUC) score to evaluate the classification performance given the imbalance classes. As shown in Table A3, a significant performance drop can be seen after removing the memory mechanism, demonstrating the effectiveness of our design.

| Methods | AUC |
|---|---|
| $SGRRG$ | 0.781 |
| $SGRRG$ w/o MEM | 0.747 |

**The threshold for anatomical location selection**: Here, we investigate the influence of the threshold $\alpha$ in anatomical location selector module which determines the anatomical locations to be included in the scene graph by varying $\alpha$ from 0.1 to 0.5. As can be seen in Figure A4, the threshold $\alpha$ plays an essential roles in $SGRRG$ where a too large or two small value both cause performance drop. Note that a higher threshold, e.g., 0.4 or 0.5, significantly impedes the performance of the model. As we mentioned earlier, radiologists may choose to ignore certain normal regions and this varies from different radiologists and samples, which however, could confuses the model trained under the cross-entropy loss. In addition, the noisy and inaccurate annotation might also lower the selection scores for the detected anatomical locations. These problems encourage us to set a lower threshold to include more anatomical locations in the scene graph and let the proposed decoder to filter out the irrelevant information. Nonetheless, an extremely low threshold, e.g., 0.1, could introduce low-quality region features into the model, thus also leading to performance drop.

## A.5 More Visualization Examples

Here, we show more visualizations of abnormal examples in Figure A3 to further verify the effectiveness of our proposed framework. Clearly, $SSGRRG$ can capture key abnormal observations in the radiology image.

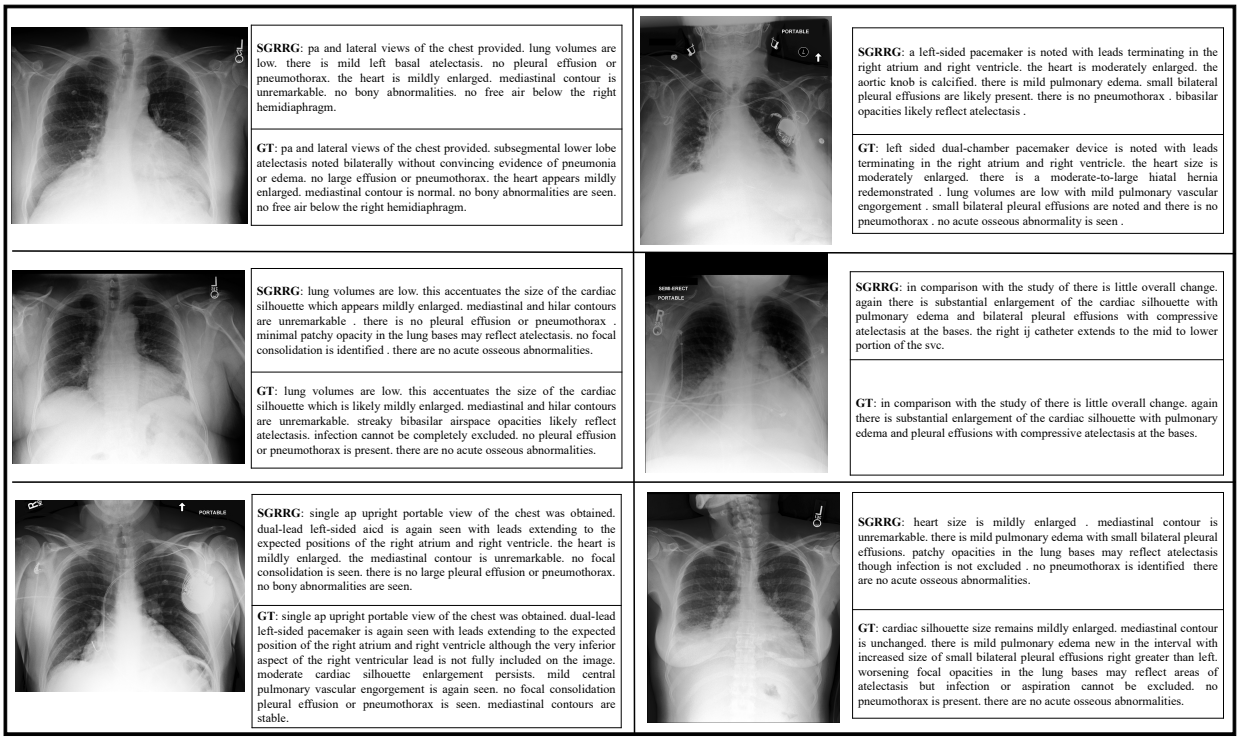

Figure A3: More visualizations of abnormal examples from the test set.

## A.6 Discussions

One may ask why we do not directly use region features or the scene graph as the only visual source, and then the visual feature extractor can be removed. Note that the bounding-box annotations in RRG are automatically generated by atlas-based bounding-box detection pipeline (not human labelled), which inevitably contain noisy and misleading supervision, and hence cannot be taken as the main visual sources. Some recent research also shows that the end-to-end architecture has better performance in image captioning. Moreover, samples without any anatomical location detected by the detector trained in these annotations cannot be processed by such a model. For example, around 5% of samples have no bounding boxes detected. In contrast, our model can still process the samples without the scene graph and benefit from both the global and local information brought by the patch-level and region-level features. The usage of the bounding box annotation to form the region-level features on-the-fly rather than an offline region features and the integration of the attribute prediction make our method be a more flexible and trained end-to-end framework.

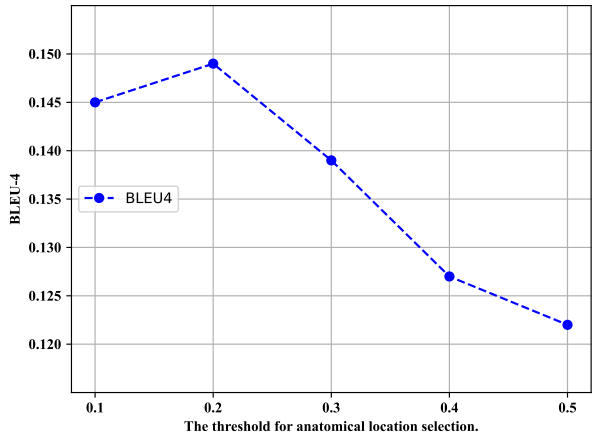

Figure A4: Ablation study for the threshold in anatomical location selector.

## A.7 Ethical Considerations

The automatic medical reporting from subject data, such as medical images, raises a number of ethical important considerations.

### A.7.1 Data Anonymity and Subjects' protected attributes

In this work we have used publicly available and anonymized data, which have become the standard benchmark datasets for radiology report generation. The data consist of gray-scale image data (JPEG/PNG) format for chest x-rays and corresponding plain-text radiology reports.

**De-identification**: For MIMIC-CXR we restate the dataset publisher's statement: "The dataset is de-identified to satisfy the US Health Insurance Portability and Accountability Act of 1996 (HIPAA) Safe Harbor requirements. Protected health information (PHI) has been removed. The dataset is intended to support a wide body of research in medicine including image understanding, natural language processing, and decision support."

### A.7.2 Translation of research findings to intended use

There are obvious societal benefits to the methods we are building, not least in reducing healthcare costs, improving the reliability and reproducibility of diagnoses from medical images and furthering the democratization of healthcare. We discuss briefly the motivations for the application and potential benefits in the introduction of our paper.

The methods we are investigating, should the performance prove adequate, would at some stage need to undergo national regulatory approvals, e.g. EMA, FDA, were they to be proposed for use in a clinic. Our work is not at this stage and we are not seeking to do this. We are also not currently aiming to validate the findings on locally collected patient data. However should the need arise, we are aware this would require approval by our medical ethics review committees. We have experience of such work on subject (clinical) data.

## References

[1] Chen, Z., Song, Y., Chang, T.H., Wan, X.: Generating radiology reports via memory-driven transformer. In: Proceedings of the 2020 Conference on Empirical Methods in Natural Language Processing (EMNLP). pp. 1439–1449 (2020)

[2] Loshchilov, I., Hutter, F.: Decoupled weight decay regularization. In: International Conference on Learning Representations

[3] Paszke, A., Gross, S., Massa, F., Lerer, A., Bradbury, J., Chanan, G., Killeen, T., Lin, Z., Gimelshein, N., Antiga, L., et al.: Pytorch: An imperative style, high-performance deep learning library. Advances in Neural Information Processing Systems **32** (2019)

[4] Tanida, T., Müller, P., Kaissis, G., Rueckert, D.: Interactive and explainable region-guided radiology report generation. In: Proceedings of the IEEE Conference on Computer Vision and Pattern Recognition. pp. 7433–7442 (2023)

[5] Wang, J., Bhalerao, A., He, Y.: Cross-modal prototype driven network for radiology report generation. In: Computer Vision–ECCV 2022: 17th European Conference, Tel Aviv, Israel, October 23–27, 2022, Proceedings, Part XXXV. pp. 563–579. Springer (2022)



\end{document}